\pdfoutput=1

\documentclass[11pt]{article}

\usepackage[final]{coling}

\usepackage{times}
\usepackage{latexsym}
\usepackage{listings}
\usepackage{enumitem}
\usepackage{algorithm}
\usepackage{amsmath}
\usepackage{algorithmic}
\usepackage[capitalize]{cleveref}
\usepackage[T1]{fontenc}

\usepackage[utf8]{inputenc}

\usepackage{microtype}
\usepackage{amsfonts}
\usepackage{inconsolata}

\usepackage{graphicx}
\usepackage{subcaption} 
\usepackage{booktabs}
\usepackage{multirow}

\usepackage{makecell}
\usepackage{array}
\usepackage{arydshln} 
\usepackage{CJKutf8}
\usepackage{bm}

\title{MBA-RAG: a Bandit Approach for Adaptive Retrieval-Augmented Generation through Question Complexity}

\author{
 \textbf{Xiaqiang Tang  \thanks{Equally contributed authors}\textsuperscript{1,2}},
 \textbf{Qiang Gao \textsuperscript{*2,3}},
 \textbf{Jian Li  \textsuperscript{*2}},
 \textbf{Nan Du\textsuperscript{2}},
\\
 \textbf{Qi Li\textsuperscript{4}},
 \textbf{Sihong Xie \thanks{Corresponding author}\textsuperscript{1}},
\\
\\
 \textsuperscript{1}The Hong Kong University of Science and Technology (Guangzhou),
 \\
 \textsuperscript{2}Tencent Hunyuan,
 \textsuperscript{3}Wuhan University,
 \textsuperscript{4}Iowa State University,
\\
 \small{
   \textbf{Correspondence:} \href{sihongxie@hkust-gz.edu.cn}{sihongxie@hkust-gz.edu.cn}
 }
}

\begin{document}
\maketitle
\begin{abstract}
Retrieval Augmented Generation (RAG)  has proven to be highly effective in boosting the generative performance of language model in knowledge-intensive tasks. However, existing RAG frameworks either indiscriminately perform retrieval or rely on rigid single-class classifiers to select retrieval methods, leading to inefficiencies and suboptimal performance across queries of varying complexity.
To address these challenges, we propose a reinforcement learning-based framework that dynamically selects the most suitable retrieval strategy based on query complexity.
Our approach leverages a multi-armed bandit algorithm, which treats each retrieval method as a distinct ``arm'' and adapts the selection process by balancing exploration and exploitation. Additionally, we introduce a dynamic reward function that balances accuracy and efficiency, penalizing methods that require more retrieval steps, even if they lead to a correct result. 
Our method achieves new state-of-the-art results on multiple single-hop and multi-hop datasets while reducing retrieval costs. Our code is available at \href{https://github.com/FUTUREEEEEE/MBA}{https://github.com/FUTUREEEEEE/MBA}
\end{abstract}


\section{Introduction and Related Work}

Retrieval Augmented Generation \cite{RAG} has shown significant promise in addressing knowledge-intensive natural language processing (NLP) tasks by integrating an updated knowledge base with a language model. This combination allows language model to access up-to-date knowledge, improving their faithfulness and reducing hallucinations.


Most existing RAG frameworks employ a retrieve-and-generate setup that indiscriminately performs retrieval based on the input. However, this approach may hinder the versatility of language models or introduce unnecessary or off-topic passages. To address this issue, methods such as SEAKR~\cite{yao2024seakrselfawareknowledgeretrieval} and FLARE~\cite{jiang2023active} have been designed to perform active retrieval only when necessary.

Furthermore, AdaptiveRAG~\cite{jeong2024adaptive} argues that real-world queries often vary in difficulty, such as the number of reasoning steps required or the depth of information needed to answer a query. Thus, applying a single retrieval method across all queries can be ineffective. Simple queries may incur unnecessary computational overhead when complex retrieval strategies are used, while complex, multi-step queries may not be adequately addressed. To tackle this issue, AdaptiveRAG introduces an adaptive router that selects the retrieval method based on the complexity of the query.
However, AdaptiveRAG simplifies the retrieval strategy into a single-choice task by using heuristic supervision that favors only one process with the least retrieval cost. This supervision is inaccurate for two main reasons. First, it assumes only one strategy is optimal for one query. For example, while multiple strategies (such as directly answering without retrieval, retrieving once, or performing iterative retrieval) might all provide correct answers depending on the query, only the direct answer method will be marked as correct. This strict supervision ignores scenarios where more complex strategies might offer better context or more comprehensive answers.
Second, this heuristic approach is ambiguous because retrieval costs vary according to the query's difficulty, what constitutes the least cost can differ significantly between queries of different complexities. 


\begin{figure*}[h!]
    \centering
     \includegraphics[width=\textwidth]{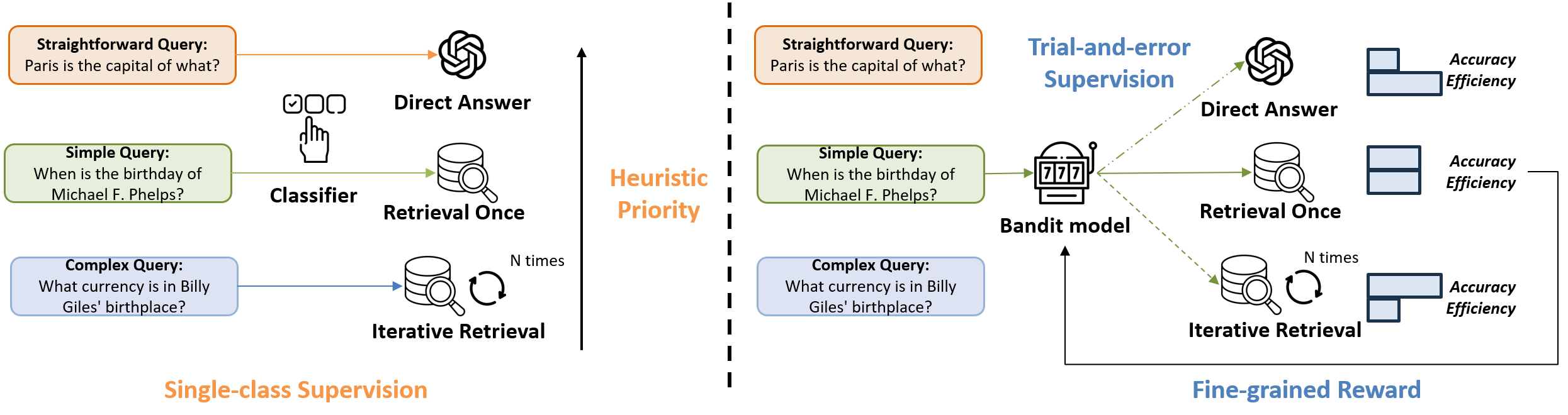}
     \vspace{-5mm}
     \caption{On the left, the AdaptiveRAG pipeline relies on an inaccurate heuristic approach to assign queries of different complexities to a single generation process. On the right, we propose an MBA-RAG framework that allows the model to explore the merits of different generation methods and rewards it based on fine-grained assessments of generation quality and cost.}
     \vspace{-5mm}
     \label{background}
\end{figure*}   

To overcome these limitations, we propose a \textbf{M}ulti-arm \textbf{B}andit-based framework for \textbf{A}daptive \textbf{R}etrieval-\textbf{A}ugmented \textbf{G}eneration (MBA-RAG) \cref{background}, that introduces both flexibility and cost-awareness into the generation process. 


First, to address the rigidity of single-label supervision, we employ a multi-armed bandit algorithm~\cite{katehakis1987multi}. The bandit framework requires only partial information for supervision; specifically, the model is supervised by the feedback of the chosen generation strategy, without indiscriminately penalizing the strategies that were not selected. This introduces greater flexibility by allowing the model to explore different generation strategies and obtain a more comprehensive selection.

Second, to tackle the issue of inaccurate heuristic supervision of retrieval costs and the varying costs associated with different generation strategies, we design a dynamic, fine-grained reward function. This function penalizes inefficient generation strategies, even if they lead to correct results, ensuring that the model not only prioritizes accuracy but also optimizes computational efficiency.

We evaluated our framework on both single-hop and multi-hop datasets. Results demonstrate that our method achieves SOTA generation accuracy while reducing retrieval costs by 20\%.

\section{Method}

Using a single retrieval method for all queries may not always be effective. In this section, we introduce a RAG framework based on a multi-arm bandit approach that dynamically selects the optimal generation method to balance generation quality and computational efficiency.

\subsection{Preliminaries}

 The RAG process \cite{RAG} includes a retrieval stage where the module \( R \) retrieves relevant documents \( D \) for a query \( x \), and a generation stage where the LLM uses \( x \) and \( D \) to generate the response \( \bar{a} = LLM(y_t|x, D) \).

\subsection{Query Encoding and Arm Selection}
\label{ArmSelectionStrategy}

To overcome the limitations of attributing each query to a single generation method, we reformulate the generation selection as a trial-and-error reinforcement learning problem, where each generation method is treated as an action (i.e. an arm) in a multi-armed bandit framework. 

We employ DistilBERT, a lightweight pre-trained language model (PLM), to encode user queries, generating an action distribution \( \mathbf{z} = f_\theta(x) \). 
Directly supervising this output could cause the system to overly favor a single generation method, leading to sub-optimal local convergence.

To prevent this, we incorporate an epsilon-greedy strategy \cite{epsilon-greedy} to balance the trade-off between exploration and exploitation \cite{UCB, auer2002finite}. This ensures that the system primarily utilizes the best-performing generation methods while still exploring other options to improve long-term performance. Specifically, the ``arm'' is chosen as \(a = \arg\max(\mathbf{z})\) with probability \(1 - \epsilon\), or a random generation method with probability \(\epsilon\).

\subsection{Learning Algorithm}

After selecting a generation method, the model updates its parameters based on both the quality and the cost of the language model's response. Unlike traditional RAG systems that prioritize accuracy alone, our approach incorporates a fine-grained reward function that accounts for both accuracy and computational cost, thereby guiding the model to minimize unnecessary computational overhead. 

The objective function, given by 
\begin{equation}
    \min_{\theta} \left( r_a - {f_\theta(x)}_a \right)^2
\end{equation}
 is to minimize the squared error between the actual reward \( r_a \) and the predicted reward \( {f_\theta(x)}_a \) for action \( a \):
\[
\theta_{t+1} = \theta_t - \alpha \nabla_\theta \left( \left( r_a - {f_\theta(x)}_a \right)^2 \right)
\]
Here, \( \theta_{t+1} \) represents the parameter set at iteration \( t+1 \), \( \theta_t \) at iteration \( t \), and \( \alpha \) is the learning rate.
The reward $r_a$ for the selected retrieval method $a$ is calculated as $r_a = \mathcal{A}(y, \hat{y}_a) - \lambda C(a)$, where $\mathcal{A}$ is a generation quality metric (e.g., exact match defined as $\mathcal{A}(y, \hat{y}_a) = \mathbb{I}\{ y = \hat{y}_a \}$ if the generated answer $\hat{y}_a$ matches the ground truth $y$)
, $C(a)$ represents the computational cost of method $a$ (such as the number of retrieval steps), and $\lambda$ is a scaling factor balancing accuracy and efficiency.




\section{Experiments}
\subsection{Datasets}
To demonstrate the performance of our method for queries of different complexities, following the work of Adaptive-RAG \citep{jeong2024adaptive}, we selected three single-hop QA and three multi-hop QA datasets as our experimental datasets.

For Single-hop QA, we choose \textbf{SquAD v1.1} \citep{rajpurkar-etal-2016-squad}, \textbf{Natural Questions} \citep{kwiatkowski-etal-2019-natural} and \textbf{TriviaQA} \citep{joshi-etal-2017-triviaqa}. These datasets consist of each query and its related articles that contain the answers. 

For Multi-hop QA, we choose  \textbf{MuSiQue} \citep{trivedi-etal-2022-musique}, \textbf{HotpotQA} \citep{yang-etal-2018-hotpotqa} and \textbf{2WikiMultiHopQA} \citep{ho-etal-2020-constructing}. For these multi-hop datasets, the answers cannot be directly obtained and require multi-step reasoning based on the information in the articles to arrive at the final answer.

\subsection{Baseline Models}
We selected a range of related models as baselines to compare with our model, including:\\
\textbf{1) No-Retrieval}: Directly generates answers without performing retrieval.
\textbf{2) Adaptive-Retrieval} \citep{mallen-etal-2023-trust}: Dynamically determining whether retrieval is necessary.
\textbf{3) Self-RAG} \citep{asai2023selfraglearningretrievegenerate}: Dynamically determines whether retrieval is needed. 
\textbf{4) DRAGIN} \citep{su-etal-2024-dragin}: evaluating the uncertainty of each token to activate the retrieval model.
\textbf{5) SEAKR} \citep{yao2024seakrselfawareknowledgeretrieval} introduce self-aware uncertainty to decide whether activating the retrieval model based on its value.
\textbf{6) Adaptive-RAG} \citep{jeong2024adaptive}: Using a classifier to dynamically choose the most suitable retrieval strategy based on the complexity of the query.

\subsection{Metrics}

Following the approach of \citet{jeong2024adaptive}, in addition to reporting performance metrics of the generated results such as Exact Match (EM), $F_1$, and Accuracy (Acc), we also report efficiency metrics of the retrieval strategy, Step.
\textbf{EM} measures whether the predicted result exactly matches the ground truth, $\bm{F_1}$ measures the overlap of words between the predicted answer and the ground truth, and \textbf{Acc} indicates whether the predicted answer contains the ground truth. \textbf{Step} denotes the number of retrieval steps required by the selected retrieval strategy.


\subsection{Experiment Settings}
We use DistilBERT \citep{sanh2020distilbertdistilledversionbert} as the query encoding model. 
This model allows for efficient query encoding and ensures that the retrieval method selection is based on a robust representation of the query.
The retrieval settings, dataset configurations, and generation model setups all follow the approach used by Adaptive-RAG \citep{jeong2024adaptive}. We utilize BM25 as the retrieval model and FLAN-T5-XL \citep{chung2022scalinginstructionfinetunedlanguagemodels} as the generation model. 
For the MBA-RAG retrieval, we employ the epsilon-greedy algorithm \citep{712192} as the action strategy.
The learning rate for the encoding model is set as 5e-5.


\begin{table}[h!]
\small
\centering
\resizebox{0.4\textwidth}{!}{
\renewcommand{\arraystretch}{1.0}
\begin{tabular}{lccccc}
\toprule

& \multicolumn{5}{c}{\bf FLAN-T5-XL (3B)}  \\
\cmidrule(l{2pt}r{2pt}){2-6} 

 \multirowcell{2}[3ex][l]{\textbf{Methods}} & EM & $F_1$ & Acc & Step   \\
\midrule


\textbf{No Retrieval$^*$} & 14.87 & 21.12 & 15.97 & 0.00 \\



\textbf{Adaptive Retrieval$^*$} & 23.87 & 32.24  & 26.73 & 0.50\\

\textbf{Self-RAG$^*$} & 9.90 & 20.79  & 31.57 & 0.72 \\

\textbf{Adaptive-RAG$^*$} & 37.17 & 46.94  & 42.10 & \underline{2.17} \\
\textbf{Adaptive-RAG (DistilBert)} & 34.37 & 43.80 & 38.50 & 1.69 \\

\textbf{MBA-RAG (Ours)(DistilBert)} & \textbf{38.80} & \textbf{48.61}  & \textbf{43.57} & \underline{1.80} \\

\textbf{MBA-RAG (Ours)(T5-Large)} & 38.40 & 48.38 & 43.30 & 1.95 \\



\bottomrule
\end{tabular}
}
\vspace{-0.1in}
\caption{Averaged results on a collection of benchmark datasets. Items marked with $^*$ have results from \citet{jeong2024adaptive}. Adaptive-RAG (DistilBert) means we use DistilBert as classifier in Adaptive-RAG settings.}

\label{tab:main}
\end{table}

\begin{table*}[h!]
\small
\centering
\resizebox{\textwidth}{!}{
\renewcommand{\arraystretch}{1}
\begin{tabular}{llccccccccccccc}
\toprule

& & \multicolumn{4}{c}{\bf SQuAD} & \multicolumn{4}{c}{\bf Natural Questions} & \multicolumn{4}{c}{\bf TriviaQA} \\
\cmidrule(l{2pt}r{2pt}){3-6} \cmidrule(l{2pt}r{2pt}){7-10} \cmidrule(l{2pt}r{2pt}){11-14}
 \multirowcell{2}[3ex][l]{\textbf{Data}} & \multirowcell{2}[3ex][l]{\textbf{Methods}} & EM & $F_1$ & Acc & Step &  EM & $F_1$ & Acc & Step  & EM &$F_1$ & Acc & Step \\

\midrule

\multirowcell{10}[-0.0ex][l]{\textbf{Single-step}} 


& \textbf{No Retrieval$^*$} & 3.60 & 10.50 & 5.00 & 0.00 & 14.20 & 19.00 & 15.60 & 0.00 & 25.00 & 31.80 & 27.00 & 0.00  \\




& \textbf{Adaptive Retrieval$^*$} & 13.40 & 23.10 & 17.60 & 0.50  & 28.20 & 36.00  & 33.00 & 0.50 & 38.40 & 46.90 & 42.60 & 0.50 \\
& \textbf{Self-RAG$^*$} & 2.20 & 11.20 & 18.40 & 0.63 & 31.40 & 39.00 & 33.60 & 0.63 &  12.80 & 29.30 & 57.00 & 0.68 \\

& \textbf{DRAGIN$^\dag$} & 18.70 & 28.70 & -- & -- &  23.20 & 33.20 & -- & -- &  54.00 & 62.30 & -- & --  \\
& \textbf{SEAKR$^\dag$}  & 27.10 & 36.50 & -- & -- &25.60   & 35.50 &  --& -- & \textbf{54.40}  & \textbf{63.10} & -- & -- \\

& \textbf{Adaptive-RAG$^*$} & {26.80} & {38.30} &{33.00} & \underline{1.37}  & {37.80} & {47.30}  & {44.60} & \underline{1.00}  &{52.20} &{60.70} & {58.20} & \underline{1.23}  \\
& \textbf{Adaptive-RAG (DistilBert)} & 22.60 & 33.90 & 28.20 & 1.57 & 34.60 & 43.00 & 39.80 & 1.58 & 49.60 & 57.50 & 54.80 & 1.32 \\

& \textbf{MBA-RAG (Ours)(DistilBert)} & \textbf{27.60} & \textbf{39.10} & \textbf{33.80} & \underline{1.11} & \textbf{37.80} & \textbf{47.50} & \textbf{44.60} & \underline{1.23} & 53.60 & 62.40 & \textbf{60.20} & \underline{1.06} \\

& \textbf{MBA-RAG (Ours)(T5-Large)} & 27.20 &19.00 & 33.40 & 1.32 & \textbf{37.80} & 47.30 & 44.55 & 1.10 & 53.20 & 62.30 & 60.00 & 1.00 \\






\midrule
\midrule

& & \multicolumn{4}{c}{\bf MuSiQue} & \multicolumn{4}{c}{\bf HotpotQA} & \multicolumn{4}{c}{\bf 2WikiMultiHopQA} \\
\cmidrule(l{2pt}r{2pt}){3-6} \cmidrule(l{2pt}r{2pt}){7-10} \cmidrule(l{2pt}r{2pt}){11-14}
\multirowcell{2}[3ex][l]{\textbf{Data}} &\multirowcell{2}[3ex][l]{\textbf{Methods}} & EM & $F_1$ & Acc & Step & EM & $F_1$ & Acc & Step & EM & $F_1$ & Acc & Step\\

\midrule

\multirowcell{10}[-0.0ex][l]{\textbf{Multi-step}} 


& \textbf{No Retrieval$^*$} & 2.40 & 10.70 & 3.20 & 0.00 &  16.60 & 22.71 & 17.20 & 0.00 & 27.40 & 32.04 & 27.80 & 0.00  \\




& \textbf{Adaptive Retrieval$^*$} & 6.40 & 15.80 & 8.00 & 0.50 &  23.60 & 32.22 & 25.00 & 0.50 &  33.20 & 39.44 & 34.20 & 0.50  \\

& \textbf{Self-RAG$^*$} & 1.60 & 8.10 & 12.00 & 0.73  & 6.80 & 17.53 & 29.60 & 0.73 & 4.60 & 19.59 & 38.80 & 0.93  \\
& \textbf{DRAGIN$^\dag$} & -- & -- & -- & -- &  23.70 & 34.20 & -- & -- &  22.40 & 30.0 & -- & --  \\
& \textbf{SEAKR$^\dag$}  & -- & -- & -- & -- &  27.90 &39.70  &  --& -- &  30.20 & 36.0 & -- & -- \\

& \textbf{Adaptive-RAG$^*$} & {23.60} & {31.80}  & {\textbf{26.00}} & \underline{3.22} & {\textbf{42.00}} & {\textbf{53.82}} & {\textbf{44.40}} & \underline{3.55}  & {40.60} & {49.75} & {46.40} & \underline{2.63}  \\





& \textbf{Adaptive-RAG (DistilBert) } & 22.20 & 30.90 & 23.80 & 2.22 & 35.80 & 47.68 & 38.00 & 1.69 & 41.40 & 49.70 & 46.40 & 1.78 \\

& \textbf{MBA-RAG (Ours)(DistilBert)} & \textbf{23.80} & \textbf{31.90} & 25.40 & \underline{2.56} &  40.60 & 52.44 & 42.60 & \underline{2.25} &   \textbf{49.40} & \textbf{58.33} & \textbf{54.60} & \underline{2.57}  \\

& \textbf{MBA-RAG (Ours)(T5-Large)} & 21.20 & 30.90 & 23.80 & 2.31 & \textbf{42.00} & 53.30 & 44.30 & 2.70 & 49.20 & 58.30 & \textbf{54.60} & 2.93 \\


\bottomrule

\end{tabular}
}
\vspace{-0.1in}
\caption{Results on each of a collection of datasets with FLAN-T5-XL (3B) as the LLM. Items marked with $^*$ have results from \citet{jeong2024adaptive}, while items marked with $^\dag$ have results from \protect\citet{yao2024seakrselfawareknowledgeretrieval}, which uses LLaMA-2-7B-Chat as the backbone LLM.
}
\label{tab:main:xl}
\vspace{-0.2in}
\end{table*}

\vspace{-0.25in}
\subsection{Main Results}
We report the average results across six datasets as shown in Table \ref{tab:main}, and the results for each individual dataset as shown in Table \ref{tab:main:xl}. Note that while we focus on performance metrics such as EM,$F_1$, and Acc, we also consider efficiency metrics like Step. The experimental results demonstrate that our method achieves a balance between the two, surpassing the baseline in performance metrics while requiring relatively fewer steps.


As shown in Table \ref{tab:main:xl}, our method has achieved performance improvements on all single-step datasets, with significant gains observed on the SQuAD and TriviaQA datasets. In particular, our approach achieves a substantial reduction in step costs, suggesting that it can dynamically select retrieval strategies with fewer retrieval steps. On the NQ dataset, our results are roughly on par with Adaptive-RAG, but with an increase in steps. This is because Adaptive-RAG, when constructing the dataset, set the condition that when zero, one, or multiple retrieval paths could yield the correct answer, the path requiring the fewest steps would be chosen as the correct label, with others marked as incorrect. However, this does not reflect reality as other paths can also yield correct answers. Therefore, our reward model does not solely choose the retrieval path with the minimum steps but may also select other feasible paths.

For the multi-step dataset, we have achieved a reduction in step costs of over 20\%, clearly demonstrating a significant enhancement in the retrieval efficiency of our method. Our method achieved significant improvements on the 2wikiMultiHopQA. On the MusiQue dataset, we also achieved results comparable to Adaptive-RAG, but with significantly lower step costs. This indicates that when two retrieval strategies can achieve comparable performance metrics, our strategy incurs lower retrieval costs. However, on the HotpotQA dataset, our method slightly underperformed Adaptive-RAG in terms of EM, Accuracy, and $F_1$, while having lower step costs. This may be because, compared to Adaptive-RAG which uses T5-large (770M) as the classifier, our method uses DistilBERT (66M), which is 10 times smaller. Unlike the other two datasets, HotpotQA is a very challenging dataset with complex question types like actoid comparison questions that demand high general knowledge performance from the model. DistilBERT's knowledge capabilities are not on par with T5. Nevertheless, our MBA-RAG still achieved more optimal strategy selection, with step costs lower than Adaptive-RAG, indicating that our MBA-RAG-based method has balanced performance metrics with efficiency metrics.

Furthermore, as shown in \cref{tab:multi-label:main}, we compare our bandit approach with a multi-label classifier. Due to the classifier's inability to consider retrieval costs, it incurs excessive computational expenses, with the majority of queries opting for a multiple retrieval strategy.
For more details, please refer to Appendix \ref{sec:appendix experiments}.

\begin{figure}[h]
    \centering
    \begin{subfigure}[b]{0.49\columnwidth}
        \includegraphics[width=\textwidth]{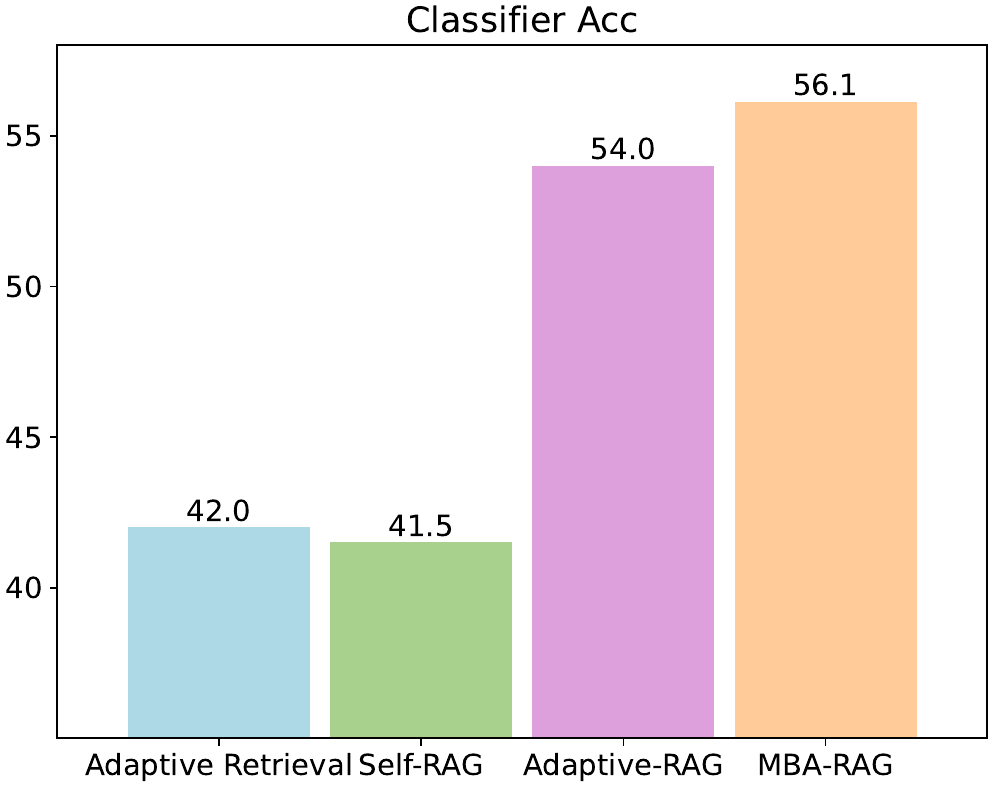} 
        \label{fig:ed}
    \end{subfigure}
    \hfill 
    \begin{subfigure}[b]{0.49\columnwidth}
        \includegraphics[width=\textwidth]{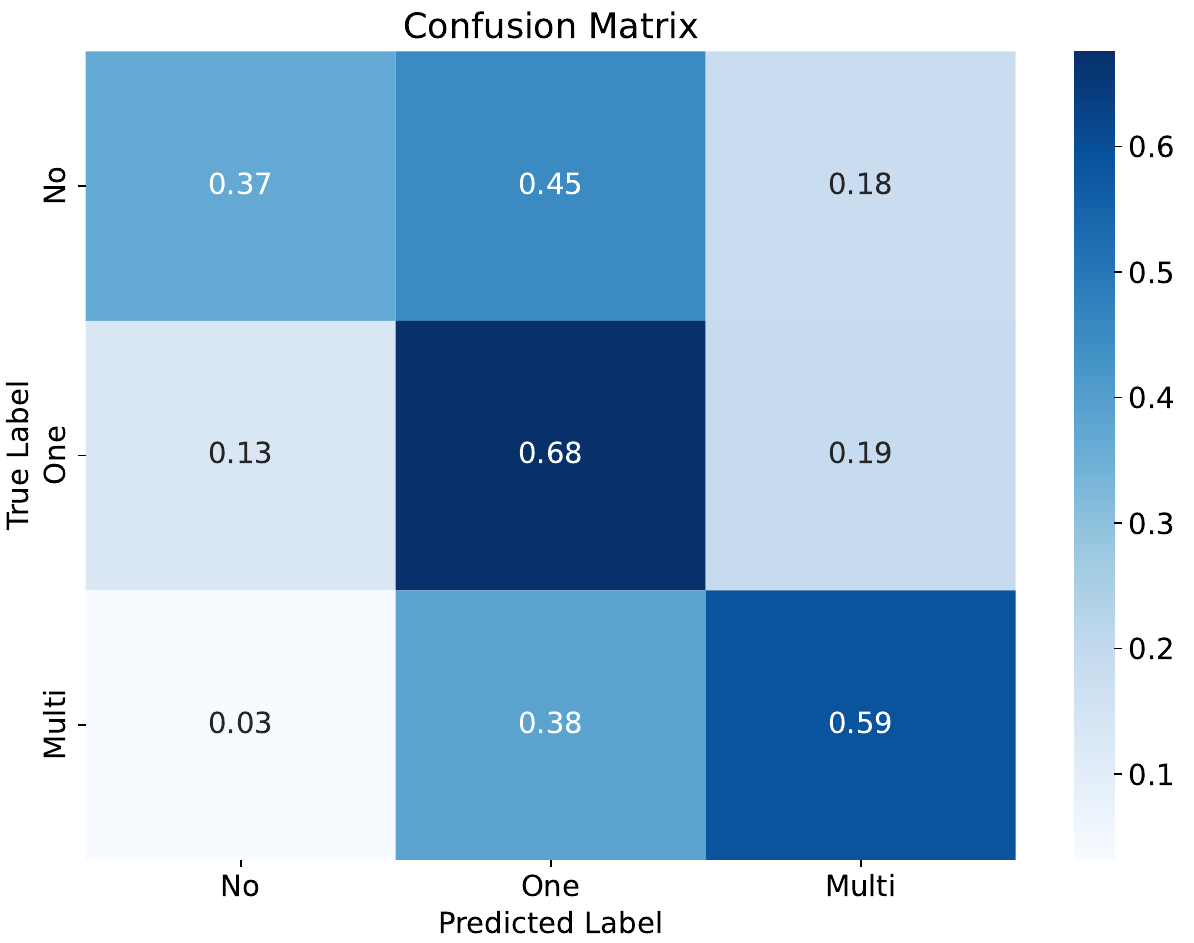} 
        \label{fig:Es}
    \end{subfigure}
\caption{On the left: classification accuracy of different adaptive approaches and the confusion matrix across all six datasets. On the right: we analyze MBA-RAG performance across different complexity labels, presenting the overall experimental results across six datasets.}
\label{fig:classifier acc}
\end{figure}

\noindent
\textbf{Retrieval Strategy Performance}
To measure the performance of different retrieval strategies, we compared the accuracy of label selection between our MBA-RAG-based classification strategy and other adaptive retrieval strategies. The results, as shown in Figure \ref{fig:classifier acc}, indicate that our classification accuracy is significantly higher than that of Adaptive-RAG. This suggests that our multi-label supervising strategy is better suited to adapt to queries of varying complexities, leading to an overall improvement in results.

\section{Conclusion}
In this study, we introduced an adaptive retrieval-augmented generation framework using a multi-armed bandit approach that dynamically selects the most efficient retrieval strategy based on query complexity. Our experimental results demonstrate the effectiveness of this method in reducing computational costs while maintaining high accuracy. Future work will aim to further refine this approach and explore its applicability to a broader range of NLP tasks.
\section*{Limitations}
Our reinforcement learning-based approach for adaptive retrieval in language models demonstrates promising results; however, it is not without limitations. The framework’s dependence on the specific structure of the multi-armed bandit algorithm can introduce challenges in scalability and adaptability to new, unseen query types. Future work could explore more efficient algorithms that maintain performance while reducing computational demands.

\bibliography{custom}

\appendix

\begin{table*}[h!]
\small
\centering
\resizebox{\textwidth}{!}{
\renewcommand{\arraystretch}{1}
\begin{tabular}{llccccccccccccc}
\toprule

& & \multicolumn{4}{c}{\bf SQuAD} & \multicolumn{4}{c}{\bf Natural Questions} & \multicolumn{4}{c}{\bf TriviaQA} \\
\cmidrule(l{2pt}r{2pt}){3-6} \cmidrule(l{2pt}r{2pt}){7-10} \cmidrule(l{2pt}r{2pt}){11-14}
 \multirowcell{2}[3ex][l]{\textbf{Data}} & \multirowcell{2}[3ex][l]{\textbf{Methods}} & EM & $F_1$ & Acc & Step &  EM & $F_1$ & Acc & Step  & EM &$F_1$ & Acc & Step \\

\midrule

\multirowcell{2}[-0.0ex][l]{\textbf{Single-step}} 

& \textbf{Adaptive-RAG} & {26.80} & {38.30} &{33.00} & \underline{1.37}  & {37.80} & {47.30}  & {44.60} & \underline{1.00}  &{52.20} &{60.70} & {58.20} & \underline{1.23}  \\
& \textbf{Multi-label-classifier (Ours)} & {24.20} & {35.40} &{29.40} & \underline{4.392}  & \textbf{38.6} & \textbf{47.70}  & {44.60} & \underline{4.514}  &{53.60} &{62.20} & {60.00} & \underline{5.152}  \\

& \textbf{MBA-RAG (Ours)} & \textbf{27.60} & \textbf{39.10} & \textbf{33.80} & \underline{1.11} & {37.80} & {47.50} & \textbf{44.60} & \underline{1.23} & \textbf{53.60} & \textbf{62.40} & \textbf{60.20} & \underline{1.06} \\

\midrule
\midrule

& & \multicolumn{4}{c}{\bf MuSiQue} & \multicolumn{4}{c}{\bf HotpotQA} & \multicolumn{4}{c}{\bf 2WikiMultiHopQA} \\
\cmidrule(l{2pt}r{2pt}){3-6} \cmidrule(l{2pt}r{2pt}){7-10} \cmidrule(l{2pt}r{2pt}){11-14}
\multirowcell{2}[3ex][l]{\textbf{Data}} &\multirowcell{2}[3ex][l]{\textbf{Methods}} & EM & $F_1$ & Acc & Step & EM & $F_1$ & Acc & Step & EM & $F_1$ & Acc & Step\\

\midrule

\multirowcell{2}[-0.0ex][l]{\textbf{Multi-step}}

& \textbf{Adaptive-RAG} & {23.60} & {31.80}  & \textbf{26.00} & \underline{3.22} & {42.00} & {53.82} & {44.40} & \underline{3.55}  & {40.60} & {49.75} & {46.40} & \underline{2.63}  \\

& \textbf{Multi-label-classifier (Ours)} & {23.00} & {31.90} & {25.80} & \underline{3.562} & \textbf{44.40} & \textbf{56.52} & \textbf{46.80} & \underline{5.32} & \textbf{49.80} & \textbf{59.05} & \textbf{55.60} & \underline{4.144} \\

& \textbf{MBA-RAG (Ours)} & \textbf{23.80} & \textbf{31.90} & 25.40 & \underline{2.56} &  40.60 & 52.44 & 42.60 & \underline{2.25} &   {49.40} & {58.33} & {54.60} & \underline{2.57}  \\

\bottomrule

\end{tabular}
}

\caption{Results of multi-label classification on each of a collection of datasets with FLAN-T5-XL (3B) as the LLM.
}
\label{tab:multi-label:main:xl}
\vspace{-0.2in}
\end{table*}
\section{Experiments Appendix}

\subsection{Implementation Details}

For the external document corpus, we use different sources depending on the dataset type:
the Wikipedia corpus preprocessed by\cite{karpukhin2020dense} for single-hop datasets, and the preprocessed corpus by \cite{trivedi2022interleaving} for multihop datasets. Approaches such as those proposed by \cite{sun2024co2,sun2024linear} can be utilized to accelerate the training process.

\label{sec:appendix experiments}
\subsection{Reward Setting}
For single-hop and multi-hop datasets, due to the differences in their tendencies for retrieval strategy selection, the single-hop dataset tends to choose the “zero” and “one” retrieval strategies, while the multi-hop dataset favors the “one” and “multiple” strategies. Therefore, for the MAB algorithm, the same set of rewards cannot achieve optimal results on both types of datasets simultaneously. Motivated by \cite{zhang2024spatial}, during our experiments, we set two different groups of reward settings. 

For the single-hop dataset, we adopted a dynamic reward approach, combining strategy selection with step costs to balance performance and efficiency,
the rewards are set at 1, 0.9, and 1-step/10 respectively for the three retrieval strategies ``zero,'' ``one,'' and ``multiple,''. When none of the three options can answer the question, the reward is set to -1. 

For the multi-hop dataset, rewards are set at 4.3, 2.3, and 1.15.

\subsection{Multi-label Classification}
Due to Adaptive-RAG defining the selection of retrieval strategies as a single-label classification problem, where multiple strategies can yield correct results, it selects the strategy with the fewest steps as the correct label, treating other choices as incorrect. This approach is somewhat unreasonable because the other choices are also correct, albeit with a higher step cost. Therefore, to make the classification choice more rational while considering step costs, we implemented a multi-label classification setup, treating all retrieval strategies that could yield the correct answer as correct labels. We trained the classification model to predict multiple potential labels and selected the most likely label during the final inference.

\subsubsection{Setting} 
In the experimental setup, for the Multi-label classifier, we maintained consistency with the Adaptive-RAG's model setting, using T5-large as the classifier and modifying the model to a multi-label classification head. The learning rate was set to 1e-4.
We employ multi-label training and use single-label form for inference, ultimately selecting retrieval generation according to the corresponding retrieval strategy. The experimental results are shown in Tables \ref{tab:multi-label:main} and \ref{tab:multi-label:main:xl}.

\begin{table}[h!]
\small
\centering
\resizebox{0.45\textwidth}{!}{
\renewcommand{\arraystretch}{1.0}
\begin{tabular}{lccccc}
\toprule

& \multicolumn{5}{c}{\bf FLAN-T5-XL (3B)}  \\
\cmidrule(l{2pt}r{2pt}){2-6} 

 \multirowcell{2}[3ex][l]{\textbf{Methods}} & EM & $F_1$ & Acc & Step   \\
\midrule

\textbf{Adaptive-RAG} & 37.17 & 46.94  & 42.10 & \underline{2.17} \\
\textbf{Multi-label-classifier (Ours)} & \textbf{38.93} &  \textbf{48.79} & \textbf{43.70} & \underline{4.514} \\
\textbf{MBA-RAG (Ours)} & {38.80} & {48.61}  & {43.57} & \underline{1.80} \\

\bottomrule

\end{tabular}
}
\caption{Averaged results of multi-label classification on a collection of benchmark datasets including single-hop and multi-hop queries.}
\vspace{-0.1in}
\label{tab:multi-label:main}
\vspace{-0.1in}

\end{table}

\subsection{Results}
The experimental results show that classifiers trained with multi-labels are highly prone to overfitting. Due to the high proportion of the ``multiple'' labels, the final model tends to exclusively choose ``multiple.'' While this achieves good performance in metrics, the step cost is excessively high, even for single-step datasets. Such choices are overly inefficient. In contrast, our MBA-RAG is essentially a form of multi-label classification, and the dynamic reward settings enable the model to make the most rational choices based on the combination of step costs, making it more practical and feasible.

Although the Adaptive-RAG-MultiLabel yields generally high results, this is because, under this training mode, the model ultimately opts for “multiple” labels. During the construction of the training set, a large number of samples are tagged with multiple labels, leading the model to completely overfit on the “multiple” label after training. However, this approach results in an excessively high number of steps, and most problems do not require multi-step retrieval.

\end{document}